\documentclass[letterpaper]{article} 
\usepackage{arxiv}

\usepackage{multirow}

\usepackage[hyphens]{url}  
\usepackage{graphicx} 

\usepackage{amsmath}
\usepackage{amssymb}

\usepackage{natbib}  
\usepackage{caption} 
\usepackage{algorithm}
\usepackage{algorithmic}

\usepackage{newfloat}
\usepackage{listings}
\DeclareCaptionStyle{ruled}{labelfont=normalfont,labelsep=colon,strut=off} 
\floatstyle{ruled}
\newfloat{listing}{tb}{lst}{}
\floatname{listing}{Listing}

\usepackage{booktabs}

\title{Taming the Implicit: Dual-Channel Risk-Aware Reinforcement Fine-Tuning for Continual Multimodal Post-Training
}
\author{
Yibei Liu\textsuperscript{\rm 2}\equalcontrib,
Jiajun Chen\textsuperscript{\rm 1}\equalcontrib,
Qianle Zhang\textsuperscript{\rm 3},
Tangyue Jin\textsuperscript{\rm 4},
Mengying Zhu\textsuperscript{\rm 1},
Meng Xi\textsuperscript{\rm 1},
Yangyang Wu\textsuperscript{\rm 1}\corresponding
}

\affiliations{
~\textsuperscript{\rm 1}Zhejiang University
~\textsuperscript{\rm 2}University of Electronic Science and Technology of China
~\textsuperscript{\rm 3}South China University of Technology
~\textsuperscript{\rm 4}China University of Geoscience
}

\begin{document}

\maketitle

\begin{abstract}
Reinforcement fine-tuning (RFT) is widely believed to inherently resist catastrophic forgetting in continual post-training of multimodal large language models. Under pronounced task distributional shifts, however, forgetting across representative RFT algorithms escalates sharply. This stems from the implicit reward-variance regularization inherent to RFT, which proves incapable of suppressing uncontrolled optimization risk. We propose \emph{Risk-Aware Policy Optimization} \textsf{(RAPO)}, the first dual-channel framework for explicit risk governance in continual RFT. On the policy channel, \emph{Risk-Aware Policy Scaling} adaptively calibrates per-sample update magnitude via rollout reliability and Fisher-inspired local predictive sensitivity; on the data channel, \emph{Risk-Aware Dynamic Bucket Sampling} reorganizes training batches through dynamic risk stratification, steering optimization toward informative yet stable samples. As a plug-and-play strategy requiring no cross-task memory, \textsf{RAPO} generalizes to any RFT algorithm without modification. On the public MLLM-CL benchmark, \textsf{RAPO} reduces final forgetting by 79.8\% relative to its RLOO backbone while retaining new-task competitiveness.

\end{abstract}

\section{Introduction}

Deployed multimodal large language models (MLLMs) ~\cite{DBLP:journals/corr/abs-2303-08774,DBLP:journals/corr/abs-2402-11684,DBLP:journals/corr/abs-2506-05453} face a fundamental operational demand that a single training run cannot satisfy: a medical imaging system must extend from chest radiograph analysis to fundus image interpretation; an enterprise assistant must continuously incorporate domain-specific expertise without eroding its general reasoning capacity. Across these settings, the common requirement is that models acquire new capabilities post-deployment, yet full retraining is prohibitive in terms of both computational cost and data accessibility. Continual Post-Training (CPT) ~\cite{DBLP:journals/csur/ZhengQSM25,DBLP:journals/corr/abs-2309-10313} addresses this requirement through sequential optimization over a stream of tasks. Sequential optimization, however, can cause catastrophic forgetting~\cite{MCCLOSKEY1989109,FRENCH1999128}, whereby learning the current task degrades acquired capabilities. Such degradation compromises the reliability of previously validated capabilities, especially in safety-critical applications.

\begin{figure}[t]
\centering
\includegraphics[width=\columnwidth]{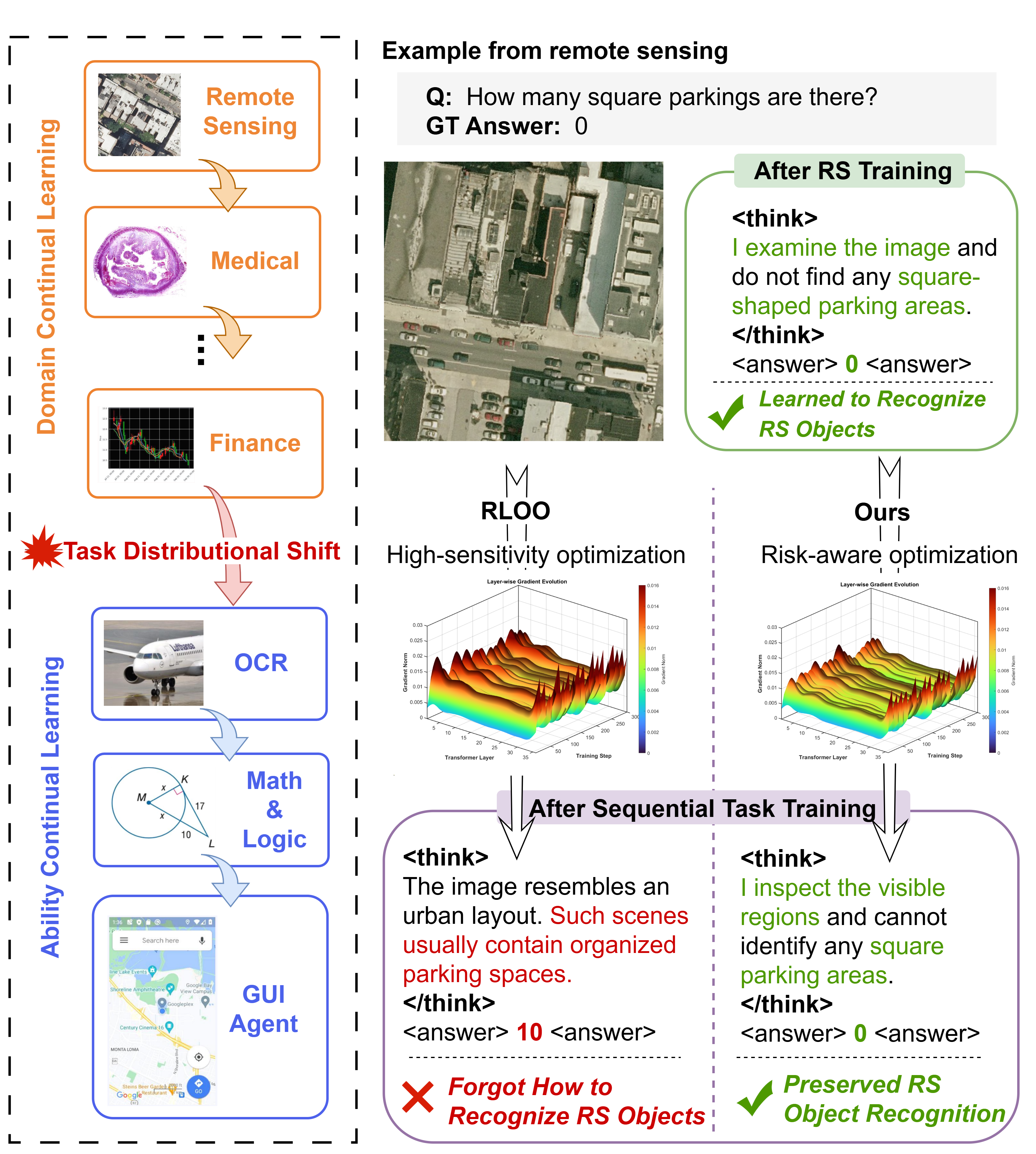} 
\caption{Continual forgetting under pronounced task distributional shifts.
After subsequent task training, RLOO forgets a previously acquired capability, whereas RAPO remains correct on the illustrated example.}
\label{fig1}
\end{figure}

Recent studies~\cite{DBLP:journals/corr/abs-2605-28860,DBLP:journals/corr/abs-2603-11653} have reported that reinforcement fine-tuning (RFT) generally retains previously acquired capabilities better than supervised fine-tuning (SFT) during CPT. One proposed explanation attributes this behavior to the implicit regularization induced by reward-dependent optimization~\cite{DBLP:journals/corr/abs-2507-05386}, making RFT a promising paradigm for continual post-training.

However, our systematic study on the MLLM-CL benchmark~\cite{DBLP:journals/corr/abs-2506-05453} shows that this retention advantage does not eliminate catastrophic forgetting. Under pronounced task distributional shifts, representative RFT methods exhibit increasingly pronounced degradation, particularly during the later Ability Continual Learning (ACL) stages. As illustrated in Figure~\ref{fig1}, standard RLOO forgets a capability acquired during an earlier task after subsequent training. This observation shows that the implicit retention of standard RFT can become insufficient under pronounced task distributional shifts and motivates explicit risk control during continual optimization.

Consequently, despite its promising properties, RFT-based continual post-training still faces three fundamental and practical challenges: (1) \emph{the absence of an RFT-oriented CPT framework}, since most existing continual-learning methods are formulated for supervised optimization rather than rollout-based policy learning; (2) \emph{limited control over implicit retention}, since the retention benefit of standard RFT varies across policy optimizers and training stages; and (3) \emph{limited adaptation of the training distribution}, since the sampling distribution in standard RFT does not respond to changes in sample-level optimization risk.

To address these challenges, we propose Risk-Aware Policy Optimization (RAPO), an explicit dual-channel risk-control framework tailored to continual RFT of MLLMs. RAPO characterizes optimization risk using rollout reliability and Fisher-inspired local predictive sensitivity. On the policy side, Risk-Aware Policy Scaling (R-Scale) converts the estimated risk into a sample-specific coefficient that adaptively scales the policy loss. On the data side, Risk-Aware Dynamic Bucket Sampling (R-Sample) converts the estimated risk into dynamic bucket assignments that adjust the composition of training batches. Both components operate on the current-stage data and require no storage of samples or task-specific statistics from previous stages.

Experiments on MLLM-CL show that RAPO reduces the magnitude of final forgetting by \(79.8\%\) relative to its RLOO backbone and achieves the highest final retained performance among the evaluated RFT methods while maintaining competitive acquisition of newly introduced tasks. Further analyses examine pronounced task distributional shifts from complementary data and optimization perspectives. They show that extreme task representational discrepancies can identify particularly vulnerable historical capabilities, while different training stages induce substantial reorganization of the layer-wise gradient structure. Ablation results demonstrate the respective contributions of R-Scale and R-Sample, and the risk-dynamics analysis verifies that RAPO adaptively adjusts policy-loss scaling and training-batch composition during continual training.

\begin{figure*}[t]
\centering
\includegraphics[width=1.0\textwidth]{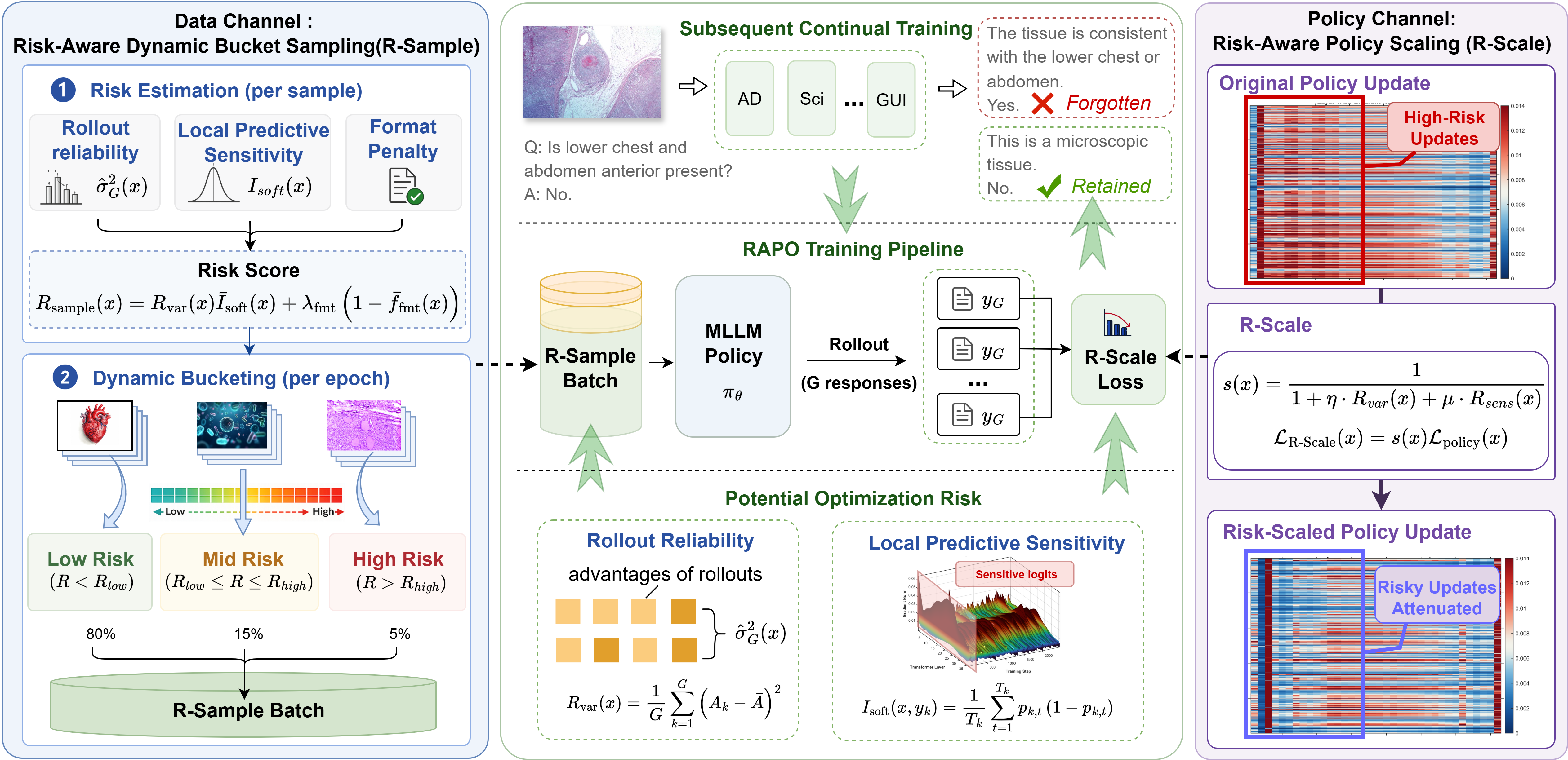} 
\caption{Overview of RAPO.
RAPO explicitly controls sample-level optimization risk through two complementary channels. R-Sample dynamically adjusts training-batch composition, while R-Scale adaptively scales the policy loss using rollout reliability and Fisher-inspired local predictive sensitivity.}
\label{fig2}
\end{figure*}

Our main contributions are summarized as follows:

\begin{itemize}
    \item We show that RFT remains vulnerable to catastrophic forgetting under pronounced task distributional shifts. We analyze this behavior through task representational discrepancy and layer-wise gradient reorganization.
    
    \item We propose RAPO, an explicit dual-channel risk-control framework tailored to continual RFT of MLLMs. RAPO characterizes optimization risk using rollout reliability and Fisher-inspired local predictive sensitivity.
    
    \item We introduce R-Scale to adaptively scale the policy loss and R-Sample to dynamically adjust the training-batch composition according to the estimated optimization risk.
    
    \item Experiments on MLLM-CL show that RAPO achieves the lowest final forgetting and the highest final retained performance among the evaluated RFT methods.
\end{itemize}

\section{Related Work}

\subsection{Continual Post-Training}
Continual Post-Training aims to continually adapt foundation models to evolving downstream tasks while preserving previously acquired knowledge across training stages. Recent MLLM continual post-training benchmarks~\cite{DBLP:journals/corr/abs-2506-05453} mainly evaluate CPT under Domain Continual Learning (DCL) and ACL, representing continual adaptation under domain shifts and capability evolution, respectively. To mitigate catastrophic forgetting, prior work has explored a variety of continual learning strategies~\cite{DBLP:conf/nips/Lopez-PazR17,DBLP:conf/iclr/MaharanaY0B25}, including Elastic Weight Consolidation (EWC)~\cite{kirkpatrick2017ewc}, Learning to Prompt (L2P)~\cite{DBLP:conf/cvpr/0002ZL0SRSPDP22}, HiDe-LLaVA (HiDe)~\cite{DBLP:conf/acl/GuoZXZWZL25}. Despite their effectiveness, these methods are almost exclusively developed under the SFT paradigm, making them difficult to directly transfer to RFT-based continual post-training. This mismatch highlights the need for CPT methods specifically designed for RFT optimization.

\subsection{Reinforcement Fine-Tuning}
Reinforcement Fine-Tuning has recently become a widely adopted paradigm for post-training foundation models by optimizing policies with reward feedback instead of supervised targets~\cite{DBLP:journals/ml/Williams92,DBLP:conf/nips/Ouyang0JAWMZASR22}. 
RFT has evolved from classical policy optimization algorithms like PPO~\cite{DBLP:journals/corr/SchulmanWDRK17} to increasingly efficient modern algorithms. GRPO~\cite{DBLP:journals/corr/abs-2402-03300} replaces value-function estimation with group-relative reward normalization to improve optimization efficiency while reducing memory consumption. RLOO~\cite{DBLP:conf/acl/AhmadianCGFKPUH24} adopts a leave-one-out baseline to reduce gradient variance~\cite{DBLP:conf/nips/GreensmithBB01} without requiring an additional value model. ReMax~\cite{DBLP:conf/icml/LiXZL00L24} further introduces a greedy-response baseline, enabling variance reduction with only a single sampled trajectory.

Recent studies have demonstrated that RFT generally exhibits stronger retention than SFT in continual post-training by substantially alleviating catastrophic forgetting while preserving general model capabilities~\cite{chen2026retaining}. Recent analyses further explain this phenomenon from different perspectives~\cite{DBLP:journals/corr/abs-2509-04259}. One line of work attributes this advantage to an implicit regularization effect, where reward variance naturally suppresses aggressive updates on uncertain samples. Another line of work explains the superior knowledge retention of RFT from the perspective of self-generated training trajectories, arguing that reinforcement learning incrementally expands existing capabilities instead of directly overwriting model behaviors~\cite{zhang2026why}. However, existing studies primarily explain why RFT mitigates forgetting, while explicit optimization risk control for continual RFT remains largely unexplored.

\section{Methodology}
\subsection{Problem Definition}

We consider RFT-based continual post-training for MLLMs. Let
$\mathcal{D}=\{\mathcal{D}_1,\mathcal{D}_2,\ldots,\mathcal{D}_T\}$
denote a sequence of training stages, each corresponding to a domain or capability to be acquired. At stage $t$, the policy $\pi_{\theta_t}$ is initialized from $\pi_{\theta_{t-1}}$ and optimized using only the current-stage dataset $\mathcal{D}_t$, without revisiting data from previous stages. The objective is to acquire the current task while preserving the capabilities learned from $\mathcal{D}_1,\ldots,\mathcal{D}_{t-1}$.

Under pronounced task distributional shifts, individual training samples may present different levels of optimization risk, while standard RFT does not explicitly regulate their contributions. We therefore formulate continual RFT as a risk-aware post-training problem that controls sample-level optimization risk during training. To instantiate this formulation, we propose RAPO, which regulates continual RFT through policy optimization and data sampling.

\subsection{Overview}
The overall framework is illustrated in Figure~\ref{fig2}. RAPO consists of a policy channel and a data channel that explicitly regulate continual RFT from complementary perspectives.

RAPO characterizes optimization risk using rollout reliability and Fisher-inspired local predictive sensitivity. Rollout reliability reflects the consistency of optimization signals across responses generated for the same input. Fisher-inspired local predictive sensitivity estimates how strongly the policy output distribution may respond to that optimization signal. Together, they characterize both the reliability of the update signal and the responsiveness of the policy, providing a more complete basis for sample-level risk control.

Based on the estimated optimization risk, the policy channel introduces R-Scale. It adaptively rescales policy optimization loss according to the estimated risk, preventing unstable updates caused by unreliable samples. Meanwhile, the data channel introduces R-Sample. It dynamically adjusts individual sample selection according to the estimated risk, encouraging stable and efficient continual learning during training. The two channels respectively regulate how strongly and how frequently individual samples contribute to continual optimization.

\begin{algorithm}[t]
\caption{RAPO Training Procedure}
\label{alg:framework}
\begin{algorithmic}[1]

\STATE \textbf{Input:} Stage datasets
\(\mathcal{D}=\{\mathcal{D}_1,\ldots,\mathcal{D}_T\}\),
\newline
initial policy \(\pi_\theta\)

\FOR{each training stage \(\mathcal{D}_t\)}

    \STATE Perform an initial rollout on \(\mathcal{D}_t\)

    \FOR{each sample \(x\in\mathcal{D}_t\)}

        \STATE Compute sampling risk \(R_{\mathrm{sample}}(x)\)

        \STATE Assign risk bucket
        \(b(x)\in\{\texttt{Low},\texttt{Medium},\texttt{High}\}\)

    \ENDFOR

    \WHILE{training on \(\mathcal{D}_t\) is not complete}

        \STATE Construct mini-batch \(\mathcal{B}\) using R-Sample

        \STATE Generate rollout responses for \(\mathcal{B}\)

        \STATE Compute rewards and response-level advantages

        \FOR{each sample \(x\in\mathcal{B}\)}

            \STATE Compute reliability risk \(R_{\mathrm{var}}(x)\)

            \STATE Compute predictive-sensitivity risk
            \(R_{\mathrm{sens}}(x)\)

            \STATE Compute R-Scale coefficient \(s(x)\)

            \STATE Update \(R_{\mathrm{sample}}(x)\) and \(b(x)\)

        \ENDFOR

        \STATE Update \(\theta\) using
        \(\mathcal{L}_{\mathrm{R\text{-}Scale}}\)

    \ENDWHILE

\ENDFOR

\STATE \textbf{Return:} Continually trained policy \(\pi_\theta\)

\end{algorithmic}
\end{algorithm}

\subsection{Risk-Aware Policy Scaling}
To explicitly regulate the optimization risk overlooked by implicit regularization, we propose R-Scale, which adaptively scales policy updates based on rollout reliability and Fisher-inspired local predictive sensitivity.

Rollout reliability characterizes the consistency of optimization signals across multiple responses generated for the same input. We quantify its corresponding risk using the variance of response-level advantages in Eq.~\eqref{eq:reliability}, where \(A_k\) is the advantage of the \(k\)-th response, \(\bar A\) is the mean advantage, and \(G\) is the number of rollout responses. A larger \(R_{\mathrm{var}}\) indicates greater disagreement among the rollout signals and therefore lower rollout reliability.
\begin{equation}
R_{\mathrm{var}}(x)
=
\frac{1}{G}
\sum_{k=1}^{G}
\left(A_k-\bar A\right)^2,
\label{eq:reliability}
\end{equation}

Fisher-inspired local predictive sensitivity characterizes how responsive the policy's output distribution is to local changes in token logits. Computing the parameter-space Fisher Information Matrix (FIM) is prohibitively expensive for large multimodal language models~\cite{DBLP:journals/neco/Amari98,DBLP:journals/jmlr/Martens20}. We therefore adopt a lightweight Fisher-inspired proxy~\cite{kirkpatrick2017ewc}, denoted by $I_{\mathrm{soft}}$ in Eq.~\eqref{eq:response_sensitivity}. For a generated token with policy probability $p$, $I_{\mathrm{soft}}$ corresponds to the associated diagonal term of the categorical Fisher matrix in logit space. It is small for saturated token probabilities and larger when the output distribution is more locally responsive. We average this token-level quantity within each response and combine it with the squared advantage in Eq.~\eqref{eq:sensitivity_risk}. A larger $R_{\mathrm{sens}}$ therefore indicates that a response simultaneously carries a stronger optimization signal and exhibits greater Fisher-inspired local predictive sensitivity during continual optimization.

\begin{equation}
I_{\mathrm{soft}}(x,y_k)
=
\frac{1}{T_k}
\sum_{t=1}^{T_k}
p_{k,t}\left(1-p_{k,t}\right),
\label{eq:response_sensitivity}
\end{equation}

\begin{equation}
R_{\mathrm{sens}}(x)
=
\frac{1}{G}
\sum_{k=1}^{G}
A_k^2
I_{\mathrm{soft}}(x,y_k).
\label{eq:sensitivity_risk}
\end{equation}

We combine rollout reliability and Fisher-inspired local predictive sensitivity into the sample-specific scaling coefficient in Eq.~\eqref{eq:rscale}. A larger estimated risk produces a smaller coefficient, thereby reducing the contribution of the corresponding sample to the policy objective.
\begin{equation}
s(x)
=
\frac{1}
{1+\eta R_{\mathrm{var}}(x)
+\mu R_{\mathrm{sens}}(x)}.
\label{eq:rscale}
\end{equation}

\begin{equation}
\mathcal{L}_{\mathrm{R\text{-}Scale}}(x)
=
s(x)\mathcal{L}_{\mathrm{policy}}(x)
+
\beta
D_{\mathrm{KL}}
\left(
\pi_\theta(\cdot|x)
\|
\pi_{\mathrm{ref}}(\cdot|x)
\right).
\notag
\end{equation}

Because R-Scale reweights the policy-loss contribution of each sample without modifying the underlying policy objective, it can be integrated with common RFT algorithms such as PPO, GRPO, RLOO, and ReMax.

\subsection{Risk-Aware Dynamic Bucket Sampling}
While R-Scale regulates policy optimization, effective CPT also requires adaptive sample scheduling. Standard RFT methods typically construct training batches without considering differences in sample-level optimization risk. We therefore propose R-Sample, which adapts the training-batch composition according to the risk estimates of individual samples throughout continual training.
\begin{equation}
R_{\mathrm{sample}}(x)
=
R_{\mathrm{var}}(x)
\bar I_{\mathrm{soft}}(x)
+
\lambda_{\mathrm{fmt}}
\left(
1-\bar f_{\mathrm{fmt}}(x)
\right).
\label{eq:sample_risk}
\end{equation}
R-Sample computes the sampling risk in Eq.~\eqref{eq:sample_risk} and assigns each sample to a low, medium, or high risk bucket according to Eq.~\eqref{eq:risk_bucket}. Here, $\bar I_{\mathrm{soft}}(x)$ and $\bar f_{\mathrm{fmt}}(x)$ denote the response-averaged predictive sensitivity and format correctness of sample $x$, respectively. The coefficient $\lambda_{\mathrm{fmt}}$ controls the format penalty, while $\tau_{\mathrm{low}}$ and $\tau_{\mathrm{high}}$ define the risk-bucket boundaries. This bucket-based organization provides a simple yet effective way to distinguish samples with different optimization priorities while avoiding hard sample selection.
\begin{equation}
b(x)
=
\begin{cases}
\mathrm{low},
&
R_{\mathrm{sample}}(x)<\tau_{\mathrm{low}},
\\
\mathrm{mid},
&
\tau_{\mathrm{low}}
\le R_{\mathrm{sample}}(x)
<\tau_{\mathrm{high}},
\\
\mathrm{high},
&
R_{\mathrm{sample}}(x)
\ge \tau_{\mathrm{high}}.
\end{cases}
\label{eq:risk_bucket}
\end{equation}

At the beginning of each training stage, an initial rollout is performed to estimate the optimization risk of all training samples and initialize the corresponding risk buckets. During training, the risk estimates and bucket assignments are updated periodically. At each sampling step, a fraction \(1-\epsilon\) of the mini-batch is drawn from the three risk buckets. Within this risk-guided portion, the sampling ratios for the low, medium, and high risk buckets are \(0.80\), \(0.15\), and \(0.05\), respectively. The remaining fraction \(\epsilon=0.10\) is sampled uniformly from the current-stage dataset. When a bucket contains insufficient samples, its remaining quota is filled through uniform sampling.

This risk-aware sampling strategy balances optimization efficiency and sample diversity while avoiding repeated selection of a small subset of training samples. As training progresses, the estimated risk is periodically updated, allowing the sampling distribution to adapt to the evolving optimization state of the policy.
\begin{table*}[t]
\centering

\small
\setlength{\tabcolsep}{7.3pt}
\renewcommand{\arraystretch}{1.18}

\begin{tabular}{llccccccccccc}
\toprule
\textbf{Category}
& \textbf{Method}
& \textbf{Med}
& \textbf{AD}
& \textbf{Sci}
& \textbf{Fin}
& \textbf{OCR}
& \textbf{Math}
& \textbf{GUI}
& \textbf{MFT}
& \textbf{MFN}
& \textbf{MAA}
& \textbf{FM}
\\
\addlinespace[2pt]
\midrule
\addlinespace[2pt]

\multirow{3}{*}{SFT}
& Seq-SFT
& -0.60
& -4.05
& -2.23
& -10.08
& -9.14
& -39.38
& -10.54
& 70.67
& 60.13
& 63.22
& -10.54
\\

& EWC
& 0.00
& -3.90
& -2.03
& -10.18
& -8.68
& -36.40
& -12.33
& 71.36
& 59.10
& 63.74
& -12.33
\\

& L2P
&-10.27
&-9.63
&-16.13
&-20.08
&-16.13
&-35.37
&-17.69
&61.63
&43.94
&42.64
&-17.69
\\

\midrule

\multirow{4}{*}{RFT}

& ReMax
& -0.30
& -5.55
& -4.23
& -2.88
& -9.44
& -11.66
& -12.74
& \textbf{64.66}
& 53.04
& \textbf{55.95}
& -12.74
\\

& GRPO
& \textbf{-0.10}
& \underline{-0.95}
& -0.47
& -0.50
& -1.70
& -5.77
& \underline{-5.96}
& 60.41
& 54.70
& 54.83
& \underline{-5.96}
\\

& RLOO
& -0.90
& \textbf{-0.35}
& \textbf{-0.07}
& \underline{-0.35}
& \underline{-1.16}
& \underline{-3.52}
& -6.53
& \underline{61.21}
& \underline{56.11}
& 55.07
& -6.53
\\

& RAPO
& \underline{-0.20}
& \textbf{-0.35}
& \underline{-0.27}
& \textbf{-0.10}
& \textbf{-0.76}
& \textbf{-0.63}
& \textbf{-1.32}
& 60.05
& \textbf{59.38}
& \underline{55.46}
& \textbf{-1.32}
\\

\bottomrule
\end{tabular}
\caption{Continual-learning performance on MLLM-CL. The stage-wise columns report FM after each training stage. Higher values are better for all metrics, and forgetting values closer to zero indicate less degradation. The best and second-best results among RFT methods are shown in bold and underlined, respectively.}
\label{tab:forgetting_comparison}

\renewcommand{\arraystretch}{1.0}
\end{table*}

\section{Experiments}
\subsection{Experimental Setup}

\textbf{Benchmark and Evaluation Protocol.}
We evaluate continual RFT on the MLLM-CL benchmark for multimodal models, which covers both DCL and ACL scenarios under a unified setting. 
Following the original task order, we construct an eight-stage continual learning sequence under this benchmark:
RS, Med, AD, Sci, Fin, OCR, Math \& Logic, and GUI Agent.
Each task contains 10K training samples.
We adopt a replay-free sequential setting, in which the model is trained on each task without accessing samples or task-specific information from previous stages.
Following the original continual benchmark, we report MFT, MFN, MAA, and FM. We primarily use MFN to characterize the overall final retained performance and FM to measure the average degradation of previously learned tasks. Since forgetting is reported as a negative performance change, a higher FM value, i.e., one closer to zero, indicates better retention.

\textbf{Baselines.}
We compare our RAPO with representative methods under both supervised and reinforcement fine-tuning paradigms. The SFT-based baselines include sequential SFT, EWC, and L2P, covering naive sequential adaptation, parameter regularization, and prompt-based continual learning. The RFT-based baselines include ReMax, GRPO, and RLOO, which employ different estimators and baselines for policy optimization. This comparison allows us to examine both the retention advantage of RFT over SFT and the additional benefit of explicit risk control over standard RFT.

\textbf{Implementation Details.}
We use Qwen2.5-VL-3B-Instruct as the base model~\cite{DBLP:journals/corr/abs-2502-13923}. Sequential SFT and all RFT methods use full-parameter fine-tuning, while EWC and L2P retain their method-specific mechanisms. All methods follow the same task sequence and training budget, with each stage trained for at least two epochs and checked for convergence. For all methods, the global batch size is set to 128 to ensure a consistent number of parameter updates. For RFT, the models are optimized with AdamW~\cite{DBLP:conf/iclr/LoshchilovH19} under BF16 precision using a learning rate of \(1\times10^{-6}\), and KL regularization is applied with a coefficient of \(1\times10^{-2}\).

For R-Scale, the coefficients $\eta$ and $\mu$ are both set to 0.25. For R-Sample, the exploration ratio $\epsilon$ is set to 0.1. The sampling ratios for low-, mid-, and high-risk buckets are set to 0.8, 0.15, and 0.05, respectively.
\subsection{Main Results}

\textbf{Overall Comparison.}
As shown in Table~\ref{tab:forgetting_comparison}, RFT-based methods generally exhibit substantially less forgetting than SFT-based methods over the complete task sequence. GRPO and RLOO provide strong retention, while ReMax remains vulnerable in this setting, and neither EWC nor L2P improves upon sequential SFT. The results show that RFT provides a favorable basis for continual post-training, although its retention advantage can weaken under pronounced task distributional shifts.

\textbf{Stage-Wise Behavior.}
GRPO and RLOO exhibit limited degradation during most DCL stages, followed by pronounced forgetting during later ACL stages. The increase is evident after Math \& Logic. Forgetting is therefore concentrated at particular stages rather than increasing uniformly throughout the sequence.

\textbf{Effectiveness of RAPO.}
RAPO achieves the lowest forgetting among evaluated RFT methods. Meanwhile, it maintains competitive final retained performance. Its retention advantage is most visible after Math \& Logic and GUI Agent, where the standard RFT methods undergo larger historical degradation. RAPO also maintains MFT and MAA values comparable to GRPO and RLOO, showing that the reduction in forgetting is not accompanied by a substantial decline in overall task acquisition.

\begin{table}[t]
\centering

\small
\setlength{\tabcolsep}{4.2pt}      
\renewcommand{\arraystretch}{1.15}
\setlength{\tabcolsep}{8pt}
\begin{tabular}{lccccc}
\toprule
Method & Med & AD & OCR & MFN & FM \\
\midrule

RLOO
& -0.90
& \underline{-0.35}
& -3.03
& 52.25
& -3.03
\\

w/o R-Scale
& \textbf{0.00}
& \textbf{0.00}
& -2.17
& 52.53
& -2.17
\\

w/o R-Sample
& \underline{-0.20}
& -0.55
& \underline{-2.03}
& \textbf{54.58}
& \underline{-2.03}
\\

RAPO
& \underline{-0.20}
& \underline{-0.35}
& \textbf{-1.13}
& \underline{53.25}
& \textbf{-1.13}
\\

\bottomrule
\end{tabular}
\caption{Ablation of RAPO over the RS-to-OCR training sequence.
Models are trained on all six stages and evaluated at the selected RS, Med, AD, and OCR checkpoints. FM values closer to zero indicate less forgetting.}
\label{tab:ablation}
\end{table}

\subsection{Ablation Study}

To examine the contribution of the two control channels, we construct two reduced variants under the same training configuration as the main experiment. The ablation follows the complete sequence from RS to OCR, where the main comparison first exhibits a clear increase in forgetting. For a focused module-level comparison, we evaluate the models at four selected checkpoints after RS, Med, AD, and OCR training.

\textit{1) Without R-Scale.}
We remove policy-loss scaling while retaining R-Sample. Training batches are therefore constructed through dynamic risk bucketing, but all selected samples are optimized using the original RFT objective.

\textit{2) Without R-Sample.}
We replace dynamic bucket sampling with uniform sampling while retaining R-Scale as the sole policy mechanism. The policy-loss contribution of each sampled instance remains regulated by rollout reliability and Fisher-inspired local predictive sensitivity.

As shown in Table~\ref{tab:ablation}, both reduced variants exhibit consistently less final forgetting than the RLOO baseline, showing that policy-loss scaling and dynamic risk-aware sampling can each improve retention when applied independently. The complete RAPO achieves the best FM, indicating that the policy and data channels provide complementary control over continual optimization.

\subsection{Task Representational Discrepancy}

We first characterize the pronounced task distributional shifts from the data perspective by examining whether the observed forgetting is associated with representational differences between the current task and historical tasks~\cite{DBLP:conf/iclr/RamaseshDR21}. Under standard GRPO, we use a frozen Qwen2.5-VL-3B model to extract multimodal representations from 500 samples per task and compute the cosine distance between task centroids. At each stage, we identify the most distant historical task and pair this maximum distance with the immediate performance change of the same task. We also report the average forgetting over all historical tasks at that stage.

\begin{figure}[t]
\centering
\includegraphics[width=\columnwidth]{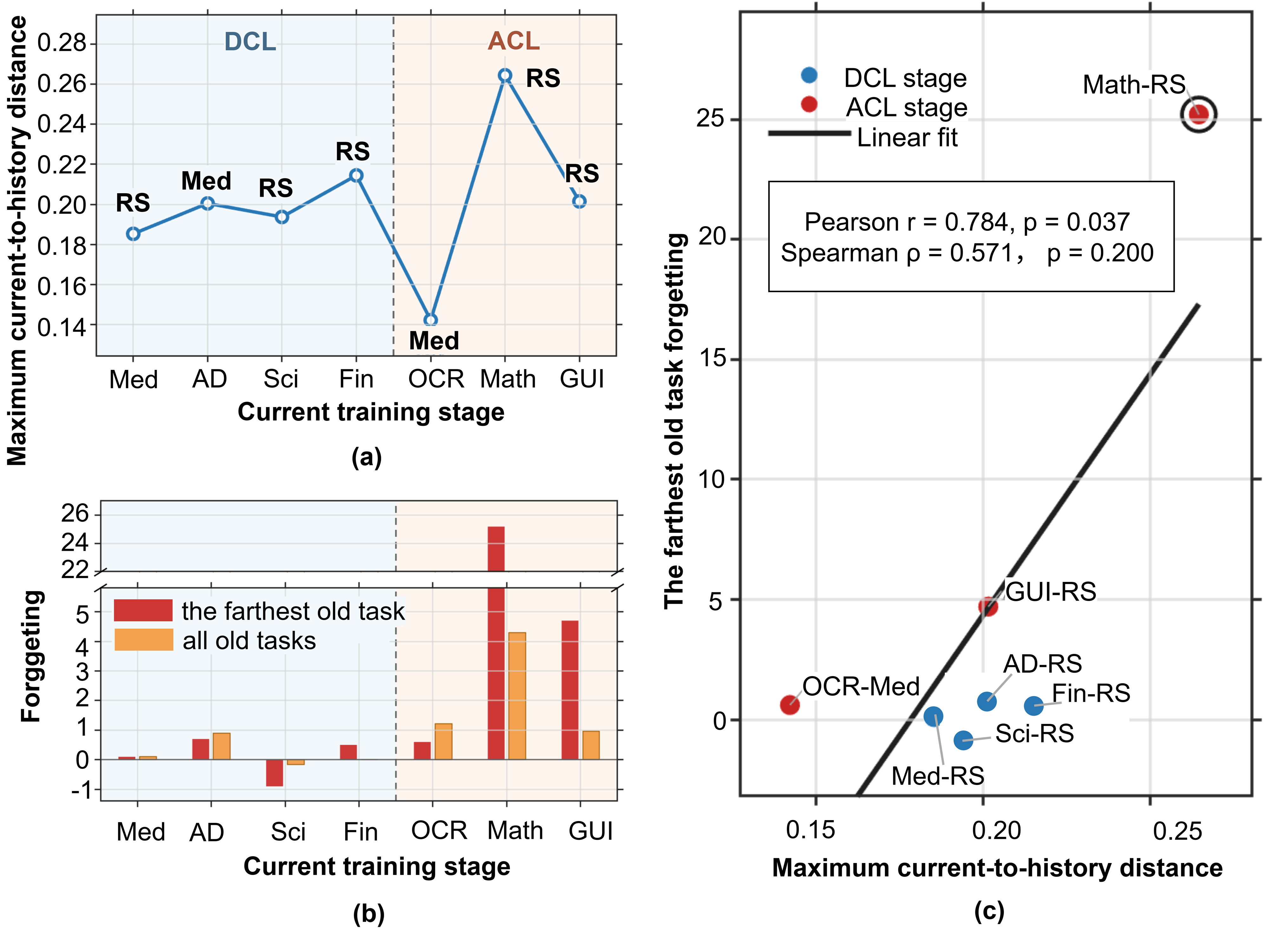} 
\caption{Task representational discrepancy and matched forgetting dynamics under GRPO.
(a) Maximum distance from the current task to its historical tasks.
(b) Forgetting of the farthest historical task and mean forgetting over all historical tasks.
(c) Relationship between maximum distance and matched forgetting.}
\label{task_distance}
\end{figure}

As shown in Figure~\ref{task_distance}, the maximum current-to-history distance remains within a relatively narrow range over the DCL stages, where matched forgetting is also limited.  The most prominent observation occurs during Math training, where RS has the largest representational distance and undergoes the most severe matched degradation. 

Figure~\ref{task_distance}(c) shows a positive association between maximum distance and matched forgetting across stages (Pearson \(r=0.784\), \(p=0.037\)). This association is largely influenced by the extreme Math--RS observation. After excluding this observation, the Pearson correlation decreases to \(r=0.179\) with \(p=0.735\), while the full-sample Spearman association is also nonsignificant. Representational discrepancy is therefore most informative for identifying extreme cases in which a historical task is particularly vulnerable to the current training stage. It does not independently explain the average forgetting of an entire stage.
\begin{figure}[t]
\centering
\includegraphics[width=\columnwidth]{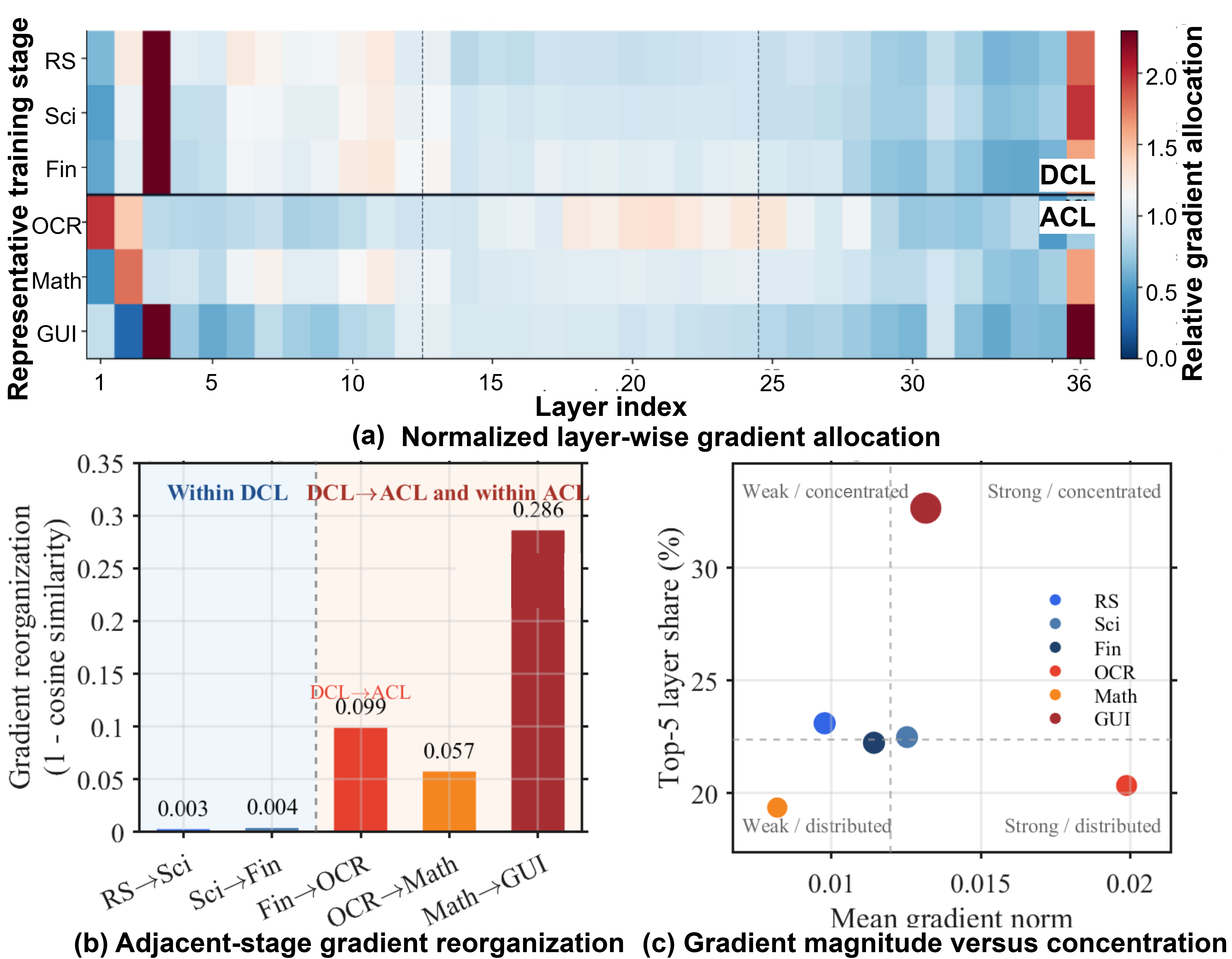} 
\caption{Layer-wise gradient reorganization under GRPO.
(a) Normalized gradient allocation across layers.
(b) Reorganization between adjacent training stages.
(c) Comparison of gradient magnitude and concentration.}
\label{gradient}
\end{figure}

\subsection{Layer-Wise Gradient Reorganization}

Task representations characterize the distribution shifts from the data perspective but do not reveal how the model responds during optimization. We therefore examine the corresponding changes in layer-wise gradient structure under standard GRPO training. We select RS, Sci, and Fin to represent the early, intermediate, and final portions of DCL, and use OCR, Math, and GUI to cover the three ACL stages. For each stage, we aggregate the layer-wise gradient norms over training steps and normalize the resulting profile. We use cosine distance to measure gradient reorganization between consecutive selected stages, while the mean gradient norm and Top-5 layer share characterize update magnitude and concentration, respectively.
\begin{figure*}[t]
    \centering
    \includegraphics[width=\textwidth]{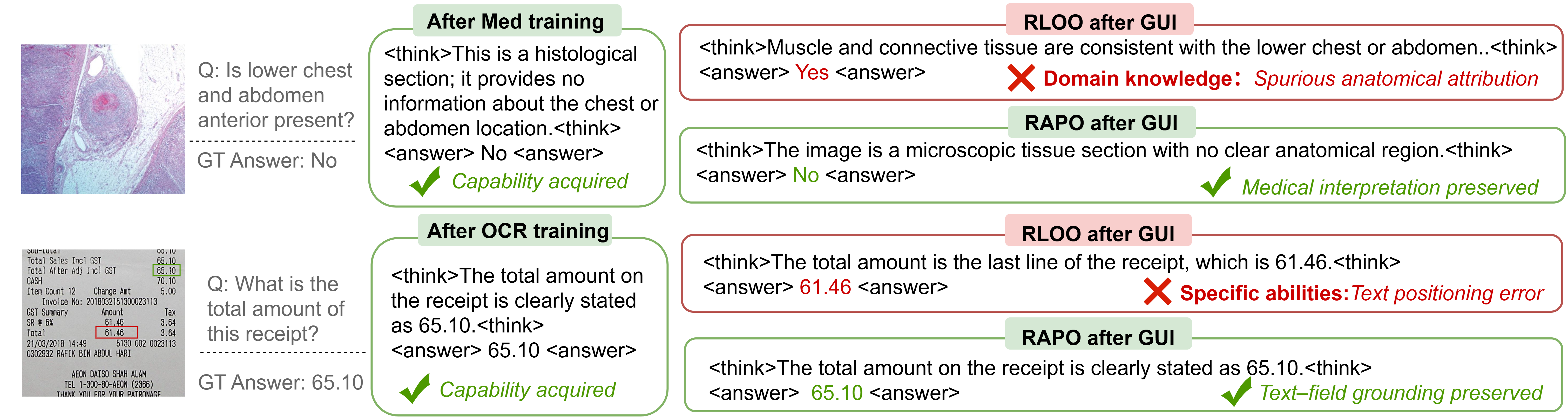}
    \caption{Qualitative comparison of capability retention.
    Both methods answer the medical and OCR examples correctly at their corresponding historical checkpoints. After subsequent GUI training, RLOO fails on domain-specific knowledge and visual text grounding, whereas RAPO remains correct on both examples.}
    \label{case_study}
\end{figure*}
\begin{figure}[t]
\centering
\includegraphics[width=\columnwidth]{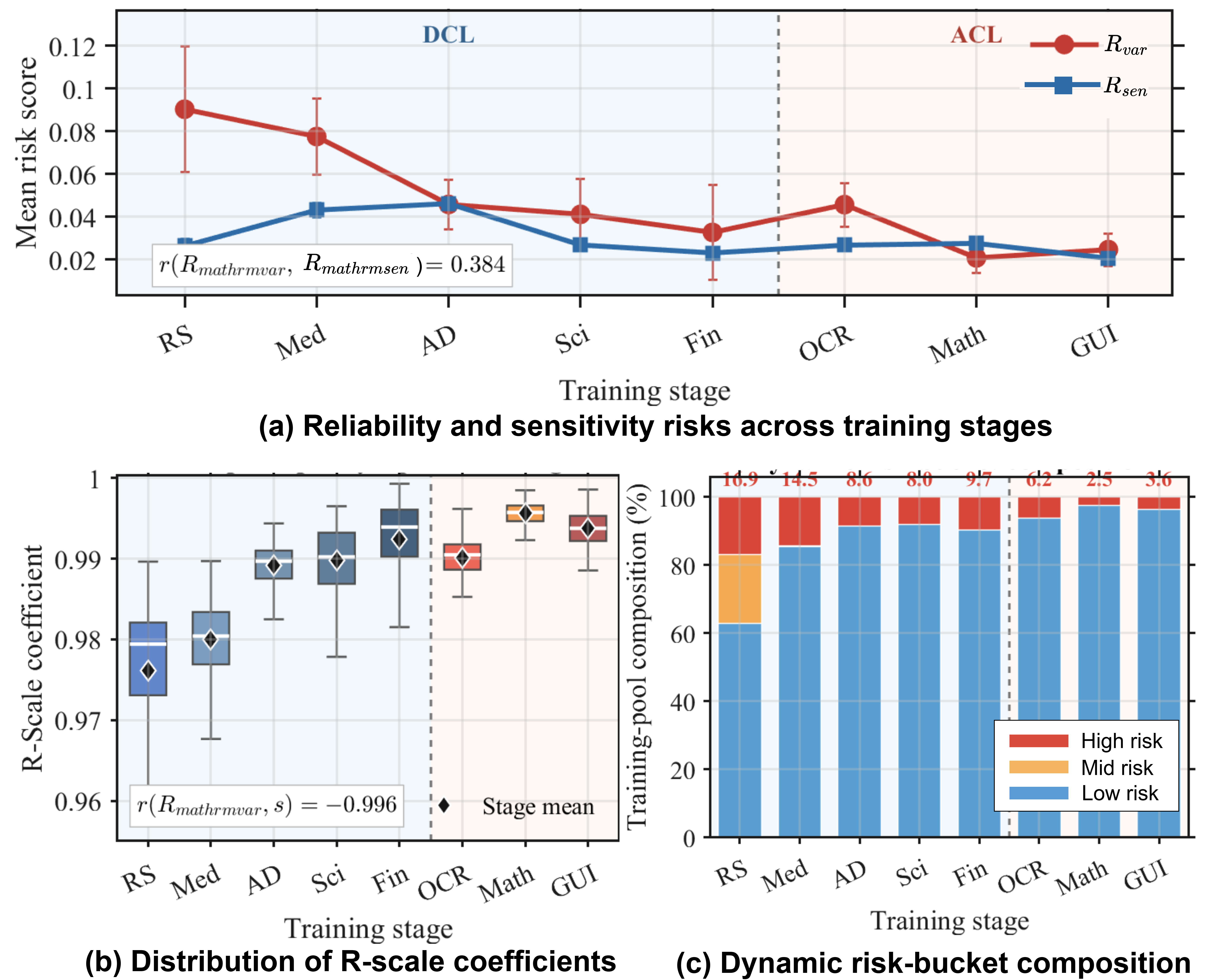} 
\caption{Risk dynamics of RAPO across training stages.
(a) Reliability and sensitivity risks.
(b) Distribution of R-Scale coefficients.
(c) Dynamic composition of the R-Sample risk buckets.
Error bars denote standard deviations, and diamonds indicate stage means.}
\label{risk_dynamics}
\end{figure}

Figure~\ref{gradient}(a) shows that the selected DCL stages maintain similar layer-wise gradient profiles, including a shared early-layer peak. Figure~\ref{gradient}(b) shows that the difference between Fin and OCR increases to \(0.099\), compared with values close to zero among the selected DCL stages. The largest reorganization occurs between Math and GUI, where the profile distance reaches \(0.286\).

Figure~\ref{gradient}(c) further shows that gradient magnitude and concentration vary independently across stages. OCR produces the largest mean gradient norm but distributes its updates relatively broadly, whereas GUI concentrates a substantially larger share of its gradient in a small number of layers. Thus, global update magnitude alone does not fully characterize how selectively an update acts across the model.

\subsection{Dynamics of Risk-Aware Optimization}

The preceding analyses characterize the pronounced task distributional shifts from the data perspective and their optimization response through layer-wise gradient reorganization. We next examine how RAPO converts its online risk estimates into policy-side and data-side controls. Across all eight stages, we record the reliability risk \(R_{\mathrm{var}}\), the Fisher-inspired local predictive sensitivity risk \(R_{\mathrm{sens}}\), and the R-Scale coefficient at every policy update. We also summarize the proportions of samples assigned by R-Sample to the low, medium, and high risk buckets.

Figure~\ref{risk_dynamics}(a) shows that \(R_{\mathrm{var}}\) generally decreases over the sequence, whereas \(R_{\mathrm{sens}}\) follows a different stage-wise trajectory. Their moderate stage-level correlation \((r=0.384)\) confirms that rollout reliability and Fisher-inspired local predictive sensitivity provide nonidentical information about the current optimization state.

Figure~\ref{risk_dynamics}(b) shows that larger reliability risk is associated with smaller R-Scale coefficients (\(r=-0.996\)). This relationship is consistent with the definition of R-Scale and verifies that larger estimated risk produces stronger policy-loss attenuation. Figure~\ref{risk_dynamics}(c) shows that the risk-bucket composition changes across training stages. This change confirms that R-Sample updates its risk partition as the policy evolves instead of assigning each sample a fixed risk category.

\subsection{Case Study}

Figure~\ref{case_study} presents representative samples that are initially answered correctly by both RLOO and RAPO but exhibit different outcomes after subsequent training. RLOO shows forgetting in general capabilities such as consistency between reasoning and final answers, domain knowledge such as recognizing specific medical targets, and specific abilities such as text recognition and field grounding. RAPO preserves the correct responses in these cases, providing additional qualitative evidence of its stronger long-term retention of previously acquired capabilities over time.

\section{Conclusion}

We studied catastrophic forgetting in continual RFT of MLLMs under pronounced task distributional shifts. Our analyses show that extreme task representational discrepancies can expose vulnerable historical capabilities, while different training stages substantially reorganize the magnitude and layer-wise concentration of policy gradients. Based on these observations, we proposed RAPO, an explicit dual-channel framework that controls sample-level optimization risk through rollout reliability and Fisher-inspired local predictive sensitivity. R-Scale adaptively regulates policy-loss contributions, while R-Sample dynamically adjusts training-batch composition. Experiments on MLLM-CL show that RAPO reduces final forgetting by \(79.8\%\) relative to its RLOO backbone while maintaining competitive acquisition of new tasks. These results demonstrate the practical value of explicit risk control for continual reinforcement fine-tuning.

\bibliography{aaai2027}

\end{document}